\pdfoutput=1
\documentclass[11pt]{article}

\usepackage[]{acl}

\usepackage{times}
\usepackage{latexsym}

\usepackage[T1]{fontenc}
\usepackage[utf8]{inputenc}

\usepackage{microtype}

\usepackage{inconsolata}

\usepackage{graphicx}

\usepackage[normalem]{ulem}
\usepackage{amsmath}
\usepackage{hyperref}
\usepackage{longtable}
\usepackage{array}

\definecolor{OliveGreen}{rgb}{0,0.6,0}
\definecolor{CornellRed}{rgb}{0.7, 0.11, 0.11}

\usepackage{booktabs, multirow, makecell}
\usepackage{enumitem}
\usepackage{xcolor,colortbl}
\usepackage{amsmath}
\usepackage{amssymb}
\usepackage{pifont}% http://ctan.org/pkg/pifont
\definecolor{green_colorblind_friendly}{HTML}{1b9e77}
\definecolor{orange_colorblind_friendly}{HTML}{d95f02}
\definecolor{purple_colorblind_friendly}{HTML}{7570b3}

\newcolumntype{C}[1]{>{\centering\arraybackslash}p{#1}}

\title{SMART: Simulated Students Aligned with Item Response Theory for Question Difficulty Prediction}

\author{
 \textbf{Alexander Scarlatos\textsuperscript{1}},
 \textbf{Nigel Fernandez\textsuperscript{1}},
 \textbf{Christopher Ormerod\textsuperscript{2}},
\\
 \textbf{Susan Lottridge\textsuperscript{2}},
 \textbf{Andrew Lan\textsuperscript{1}}
\\
 \textsuperscript{1}University of Massachusetts Amherst,
 \textsuperscript{2}Cambium Assessment
\\
 \small{
   \texttt{\{ajscarlatos,nigel,andrewlan\}@cs.umass.edu}
   }
\\
 \small{
   \texttt{\{christopher.ormerod,susan.lottridge\}@cambiumassessment.com}
 }
}

\begin{document}
\maketitle
\begin{abstract}
Item (question) difficulties play a crucial role in educational assessments, enabling accurate and efficient assessment of student abilities and personalization to maximize learning outcomes. Traditionally, estimating item difficulties can be costly, requiring real students to respond to items, followed by fitting an item response theory (IRT) model to get difficulty estimates. This approach cannot be applied to the cold-start setting for previously unseen items either. In this work, we present SMART (\textbf{S}i\textbf{m}ulated Students \textbf{A}ligned with I\textbf{RT}), a novel method for aligning simulated students with instructed ability, which can then be used in simulations to predict the difficulty of \emph{open-ended} items. We achieve this alignment using direct preference optimization (DPO), where we form preference pairs based on how likely responses are under a ground-truth IRT model. We perform a simulation by generating thousands of responses, evaluating them with a large language model (LLM)-based scoring model, and fit the resulting data to an IRT model to obtain item difficulty estimates. Through extensive experiments on two real-world student response datasets, we show that \mbox{SMART} outperforms other item difficulty prediction methods by leveraging its improved ability alignment.
\end{abstract}

% SECTION

\section{Introduction}

% \iffalse
% \begin{table}
% \small
% \centering
% \begin{tabular}{p{0.95\linewidth}}

% \toprule

% \textbf{Skills:} Reading Comprehension Argumentation\\
% \textbf{Grade:} 6\\
% \textbf{Passage Length:} $5864$ tokens, $X$ paragraphs\\
% \textbf{Passage:}
% \textcolor{purple_colorblind_friendly}{[paragraph 5]} \ldots The two mid-term exams are not cumulative 
% \ldots The final exam is cumulative \ldots\\
% Exam Replacement: If you score higher on the final exam than on one of your mid-term exams, the final exam score will replace that mid-term exam score. This is only true if you take both.
% \\
% \ldots\\
% \textcolor{purple_colorblind_friendly}{[paragraph 9, 10]} \ldots Topics that will be covered in the order shown below and we will spend roughly a week on each topic. 1. The cell \ldots\ Detailed topic list 1) The Cell a) Organization b) membranes \ldots\\
% \midrule
% \textbf{Rubric:} For score 0 observe\\
% \ldots\\
% For score 1 observe\\
% For score 2 observe\\
% \midrule
% \textbf{Question:} If I don't do very well on one of my midterms, will I be able to pull up my grade somehow?\\

% \bottomrule
% \end{tabular}
% \caption{Example item (passage, question, rubric). \nsf{@al}}
% \label{tab:dataset}
% \end{table}
% \fi

Assessing student knowledge and skill levels, an important aspect of education, requires carefully designed assessment items (i.e., questions) with high quality and validity. According to the Standards for Educational and Psychological Testing~\cite{eignor2013standards}, various factors play a role in the quality and validity of items, with a crucial factor being their difficulty, i.e., how challenging they are to real students. Item difficulties also play a critical role in learning platforms, enabling personalized learning through student-tailored item recommendation~\cite{chen2008personalized,ueno2019uniform}, item generation~\cite{jiao2023automatic,kurdi2020systematic,shimmei2022automatic,ashok-kumar-etal-2023-improving}, and curriculum design~\cite{mehrens1987sensitivity,zhang2024item} to maximize learning outcomes.
%Intuitively, question difficulty is a numeric measure quantifying how challenging a question is to a student.

Since new items are constantly developed and used in practice, estimating their difficulty before assigning them to students is crucial. Traditionally, estimating item difficulties can be costly and time-consuming~\cite{alkhuzaey2024text}: we need (possibly a large number of) real students to respond to them, and then fit an item response theory (IRT) model on their responses to get item difficulty estimates. In contrast, recent work has estimated item difficulties from their (mostly textual) content~\cite{tack-etal-2024-itec,gombert-etal-2024-predicting,benedetto2023quantitative,tacnn}, with some methods leveraging large language models (LLMs)~\cite{he2024psychometric,duenas-etal-2024-upn}. 
%\ml{drop cites here} 
This approach works well for many items, especially multiple-choice questions (MCQs). Among them, an interesting approach is to use \emph{simulated students}~\cite{he2024psychometric,feng2025reasoningsamplingaugmentedmcqdifficulty,simstudentsmarter} to generate responses to new items at scale, to estimate their difficulties. Initial work in this direction uses simulated students to model student responses to MCQs, followed by aggregating their predicted responses and obtaining item difficulty estimates.

\begin{figure*}
\centering
\includegraphics[width=\linewidth,trim={0.8cm 0 0.6cm 0},clip]{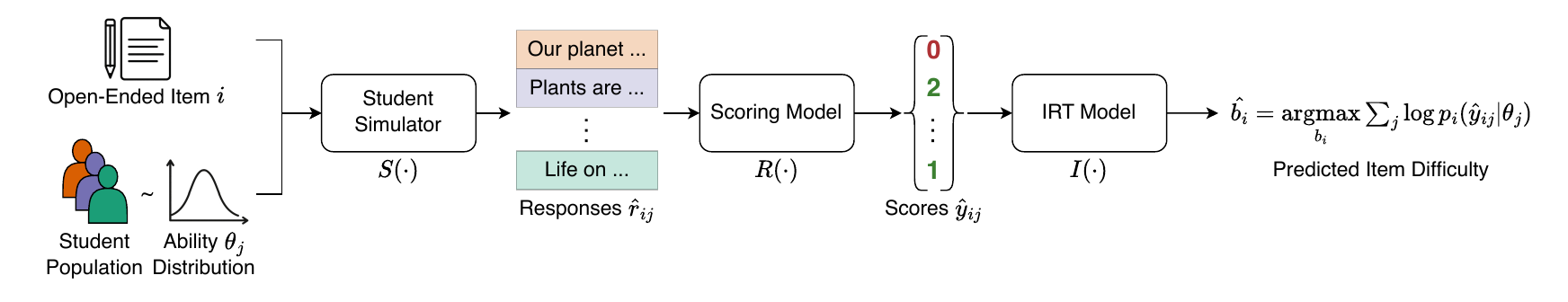}
\caption{Our three-stage pipeline for item difficulty prediction: simulate open-ended responses for students with different abilities, score them, and fit an IRT model on the scores to obtain difficulty estimates.}
\label{fig:model}
\end{figure*}

Among different item types, open-ended items require students to write short answers, and are common in many domains including math, programming, and particularly in reading comprehension assessment; see Table~\ref{tab:qual_ex_sb} for an example. Since these items require a student to construct an entire answer, they may reveal deeper insights into student knowledge~\cite{brown,feldman,smith,anderson} than other item types, such as MCQs. However, estimating the difficulties of open-ended items can be more challenging, since scoring open-ended responses accurately at scale is not as easy as for MCQs. Moreover, there is considerable diversity in terms of both response content and style among real students, which means that it can be difficult to simulate student responses to these items~\cite{sonkar-etal-2024-student}.
%Researchers have found evidence that students' open-ended responses contain useful information on their knowledge states by revealing misconceptions~\cite{?} and gaps in their knowledge~\cite{?}.
%However, there is limited existing work exploring the applicability of student simulation-based difficulty prediction of open-ended short-answer items. 

%Traditional field-testing methods are often resource and time-intensive~\cite{?} and can lead to item exposure concerns~\cite{?}.
%Recent advances in AI have led to the rise of 

%Unlike MCQs, student simulation for open-ended items is challenging requiring the generation of exact student responses, which have a high diversity in content and style. Further, for accurate IRT calibration of item difficulties, the simulated student responses have to be precisely aligned with input student abilities and question difficulties.

% SUBSECTION

\subsection{Contributions}

% In this paper, we propose SMART, a novel, three-stage pipeline, which simulates student responses to short-answer, open-ended items, to help estimate their difficulty.

In this paper, we present SMART (\textbf{S}i\textbf{m}ulated Students \textbf{A}ligned with Item \textbf{R}esponse \textbf{T}heory), a pipeline for training LLMs as simulated students that generate open-ended responses, aligning them with an instructed ability level and item difficulty.

First, SMART leverages generative LLMs as the student simulator to synthesize \emph{a distribution of} student responses to open-ended items. These responses provide valuable \textit{interpretability} in the difficulty estimation process. We use direct preference optimization (DPO) to train our student simulator, using a novel method to form preference pairs. We \textit{prefer student responses with higher likelihood under an IRT model}, thus aligning the generated responses with both student ability and item difficulty. Second, we use an LLM-based scoring model to score the generated responses. Finally, we train an IRT model on the scored student responses to obtain item difficulty estimates.

We conduct extensive quantitative experiments on two real-world datasets, one with 49 short-answer reading comprehension items and another with 50 Java coding items. We find that SMART, by simulating IRT-aligned responses to unseen items from a population of students with different abilities, outperforms state-of-the-art item difficulty prediction methods, even in the low-resource setting with few training items.
We analyze the simulated student responses and find that SMART responses correlate strongly with their instructed ability, and that the feature distributions of its responses match closely with the ground-truth. We also qualitatively compare simulated student responses across methods, perform a failure pattern analysis, and suggest directions for future work.

% \begin{enumerate}[noitemsep,topsep=0.2pt]
%     \item First, SMART leverages generative LLMs as the student simulator to synthesize \emph{a distribution of} student responses to open-ended items. These responses provide valuable \textit{interpretability} in the difficulty estimation process. We use direct preference optimization (DPO) to train our student simulator, using a novel method to form preference pairs. We \textit{prefer student responses with higher likelihood under an IRT model}, thus aligning the generated responses with both student ability and item difficulty. Second, we use an LLM-based scoring model to score the generated responses. Finally, we train an IRT model on the scored student responses to obtain item difficulty estimates.
%     \item We conduct extensive quantitative experiments on two real-world datasets, one with 49 short-answer reading comprehension items and another with 50 Java coding items. We find that SMART, by simulating IRT-aligned responses to unseen items from a population of students with different abilities, outperforms state-of-the-art item difficulty prediction methods, even in the low-resource setting with few training items.
%     \item We analyze the simulated student responses and find that SMART responses correlate strongly with their instructed ability, and that the feature distributions of its responses match closely with the ground-truth. We also qualitatively compare simulated student responses across methods, perform a failure pattern analysis, and suggest directions for future work.
% \end{enumerate}

% SECTION

\section{Background: Item Response Theory}

Item response theory (IRT)~\cite{rasch1960studies} is a framework for jointly estimating scalar-valued abilities of students and parameters of ``items'' (or questions), mainly difficulty and discrimination, from student responses to items. In the simplest case,  each student is represented by a single scalar ability $\theta_j \in \mathbb{R}$, each item is represented by a single scalar difficulty $b_i \in \mathbb{R}$, and each score $y_{ij}$ is binary, i.e, 0 (incorrect) or 1 (correct). This simple model is known as the 1-parameter logistic (1PL) model \cite{rasch1960studies}, and uses the following formulation:
\begin{align}
    P^\text{1PL}_i(y_{ij}|\theta_j) = \sigma(\theta_j - b_i). 
\end{align}
%where $\theta_j$ is the learnable ability for student $j$, $b_i$ is the learnable difficulty for item $i$, and $y_{ij}$ is the binary response that student $j$ gave on item $i$. 
In practice, 
%it is common to assume that 
$\theta_j$ and $b_i$ 
% \ml{is difficulty assumed to be standard normal too?}
% \nsf{not sure, but this says normal prior is used for theta and diff: \url{https://mc-stan.org/docs/2_28/stan-users-guide/item-response-models.html\#pl-rasch-model}}\nsf{@al}
are assumed to be drawn from a standard normal, with implementations using expectation maximization or variational inference \cite{lalor-etal-2019-learning} to enforce standard normal priors on these variables.

In our setting, we study \emph{short-answer} questions where the response is open-ended. We assume that students can receive partial credit on responses, leading to ordinal or continuous scores. For ordinal response data, we use the generalized partial credit model (GPCM) \cite{muraki1992generalized}, which adds a ``step'' parameter for each of the $C$ total score categories:
\begin{align}
    \label{eq:gpcm}
    P^\text{GPCM}_i(y_{ij}|\theta_j) = \frac{e^{\sum_{y=0}^{y_{ij}}  a_i (\theta_j - b_i + d_{iy})}}{\sum_{c=0}^{C-1}e^{\sum_{y=0}^{c} a_i (\theta_j - b_i + d_{iy})}},
\end{align}
where the response $y_{ij} \in \{0, \ldots, C-1\}$, $d_{iy}$ is the step parameter for item $i$ for score $y$, and $a_i$ is the discrimination parameter for item $i$. GPCM smoothly adjusts the probability of different ordinal scores according to student ability and item difficulty, with step parameters determining where the probability mass shifts from one score to the next.

We also handle continuous scores, i.e., $y_{ij} \in [0, 1]$, where the student's response lies on a spectrum between incorrect and correct. In this setting, we model correctness probability using a continuous Bernoulli distribution \cite{loaiza2019continuous}:
\begin{align}
    \label{eq:bernoulli}
    P^\text{Bern.}_i(y_{ij}|\theta_j) = &P^\text{1PL}_i(Y=1|\theta_j)^{y_{ij}} \cdot \\ 
    \nonumber &P^\text{1PL}_i(Y=0|\theta_j)^{(1 - y_{ij})}.
\end{align}

% SECTION

\section{Problem Formulation}

We assume that we have access to a real-world dataset of human-scored student responses to open-ended short-answer items.
%, assessing argumentation, explanation, and narration in reading comprehension. 
%\ml{not sure we need to get into the subject details here, before we detail the dataset. making things as generic as possible is good} 
Each item, indexed by $i$, is associated with 1) a short question text $q_i$, 2) (optional) meta data $m_i$, e.g., a passage text for reading comprehension items, 3) (optional) a scoring rubric $w_i$, and 4) a set of human-scored student responses denoted by $D_i = \{r_{ij}, y_{ij}\}$. An IRT model is fit to this data to get ground-truth estimates of each student ability $\theta_j$ and item difficulty $b_i$. Our goal is to predict the item difficulty parameters of previously \emph{unseen} test items before students respond to them.

% SECTION

\section{Methodology}

%Although our dataset contains a limited number of items, \ml{ok so i don't think we should emphasize this point; our method applies when there's a lot of items too, right? it's good to make it sound generic, and then after introducing the dataset, discuss what's special about this setting and what makes our method such a great fit} a key observation is our access to a \textit{large bank of human-scored student responses} ($85$K) across items, with an average of $X$ responses per item. 

Our approach, SMART, illustrated in Figure~\ref{fig:model}, uses a novel, three-stage pipeline, which simulates student responses to short-answer, open-ended items, to help estimate their difficulty.
First, SMART leverages human-scored responses by real students to \textit{open-ended items} to learn a student response simulator, $S(\cdot)$, parameterized by a generative LLM. The simulator $S(\cdot)$ generates a \textit{distribution of student responses} to unseen open-ended items. Second, an LLM-based scoring model, $R(\cdot)$, scores these generated responses.
Third, an IRT model, $I(\cdot)$, is trained on the scored student responses to estimate item difficulties.
%\ml{i would suggest using something like $R(\cdot)$, $S(\cdot)$, etc. to denote these models. will make things more clear}\nsf{\cmark}
%Unlike existing work in student simulation~\cite{feng2025reasoningsamplingaugmentedmcqdifficulty,simstudentsmarter,?} which focuses on close-ended items like MCQs, SMART, leverages an ability-aligned generative LLM to simulate short answers to open-ended items. Further, SMART trains a rubric-based scoring model to automatically score these generated responses for further IRT fitting, unlike MCQ responses which can be easily scored. 

%See Figure~\ref{fig:model}. Describe overall algorithm (need a figure for this) - 1) fit IRT model on train data, 2) use those parameters as needed to fit student simulator, 3) generate simulated responses on test set, 4) automatically score those responses, 5) re-fit IRT model on both train and test where the test set now has only simulated responses (train set needs to be included here so calibration is consistent). Prior works have done similar versions of this, key difference is how we do (2) and that (4) is needed since other works don't do open-ended items.
%Explain each component in subsequent subsections.

% SUBSECTION

\subsection{LLM-based Student Response Simulation}
\label{subsec:responsegensft}

%General setup - LLM-based simulated students, ability and stimulus (and possibly passage) as input, open-ended response as output. Ability buckets based on IRT $\theta$. Initially do SFT based on groun-truth responses.

% We first detail our LLM finetuning method for student simulation. We use a generative model from the Llama 3~\cite{grattafiori2024llama} family, specifically \texttt{Llama 3.2-3B-Instruct}, as our base LLM to parameterize our student simulator $S(\cdot)$.
We first detail our LLM finetuning method for student simulation, resulting in a model we denote as $S(\cdot)$.
%\nsf{, since smaller models have been shown to be effective student simulators~\cite{?}.} 
%\ml{rly? there's evidence for that?} 
%\nsf{have to find, they usually use smaller models as students, since maybe these models can make mistakes more easily? trying to get a reason for llama-3B}
% We combine the passage text $p_i$, question text $q_i$, and the student ability $\theta_j$ in a carefully designed template, with instructions on how to simulate student responses, to form our input prompt, and maximize the likelihood of the ground-truth student response $r_{ij}$ at output.
We combine the question text $q_i$, the meta data text $m_i$, and the student ability $\theta_j$ in our input prompt, and train the LLM using supervised finetuning (SFT) to maximize the likelihood of the ground-truth student response $r_{ij}$ as output. We convert $\theta_j$ to a string after rounding to 4 decimal places, which we find works better than binning values into descriptive buckets; we leave an exploration of more advanced methods for numeric inputs to LLMs, such as weighting model parameters \cite{wang-etal-2024-conditional}, for future work. See the exact prompt format in Appendix~\ref{sec:prompts}. 

% To include the student ability $\theta_j$, we explore two ways: either simply using the raw value, or binning the abilities into $10$ descriptive buckets described in Appendix~\ref{?}\nsf{@al}.
%, following $X$ standard nomenclature~\cite{?}. 
%\ml{what is this last part? i have no idea lol}
%, and a combination of both the raw value and the bucket description. 
%\al{I'm actually not using the buckets since they lead to lower performance, mainly because after fitting the new IRT parameters the distribution shifts downwards so the higher buckets aren't represented at all} 
%\nsf{even worse than bucket+value, right?} \al{never tried this, I could see it going either way}
% We empirically find that using the raw real-valued ability values leads to better performance, possibly due to the imprecise nature of ability bucket descriptions.  
% Although prior work has explored LLM finetuning from question text to response text for MCQs~\cite{feng2025reasoningsamplingaugmentedmcqdifficulty,simstudentsmarter,?}, our addition of student ability $\theta_j$ is key for ability-controlled student response generation for open-ended items. Responses aligned with student abilities promote better IRT fitting leading to accurate estimates of test item difficulties.

% SUBSECTION

\subsection{SMART: Ability Alignment in Response Simulation via Preference Optimization}

%Goal is to fit simulated student output distribution to ground-truth IRT parameters for ability-following and item difficulty calibration. Use DPO to achieve this.

We now motivate and detail SMART, our novel method aligning simulated student responses with their given ability and the expected difficulty of the given item.
% for promoting simulated student responses to more closely resemble the desired ability values. 
%Finetuning the LLM-based student simulator from (passage, question, ability) triples to ground-truth student responses helps induce ability-guided response generation. 
Even though the LLM-based student simulator $S(\cdot)$ is finetuned to maximize the likelihood of the textual content of student responses,
% we observe that student responses have high diversity in content and style, even among students with similar ability values. This diversity makes it difficult to align generated responses with desired ability values using only fine-tuning. 
we observe that the responses it generates do not always align well with the prompted ability value, likely because the SFT objective treats all output sequences equally without any special attention paid to the prompted ability.
%While a finetuned LLM-based student simulator serves as a good initialization, the generated responses could be noisy with respect to the input student ability leading to suboptimal IRT fitting on test items.

To address this limitation, we propose the following idea: to incorporate information on the IRT likelihoods of student responses, in addition to their textual content. The intuition is that generating responses with high likelihoods under the IRT model should lead to better alignment between the IRT parameters and response content, which should benefit the downstream task of item difficulty prediction.
To achieve this alignment, we use DPO~\cite{rafailov2023direct} to \textit{prefer responses with scores having a higher IRT likelihood} with the input student ability and item difficulty.

\begin{figure}
\centering
\includegraphics[width=\linewidth,trim={0.8cm 0 0.6cm 0},clip]{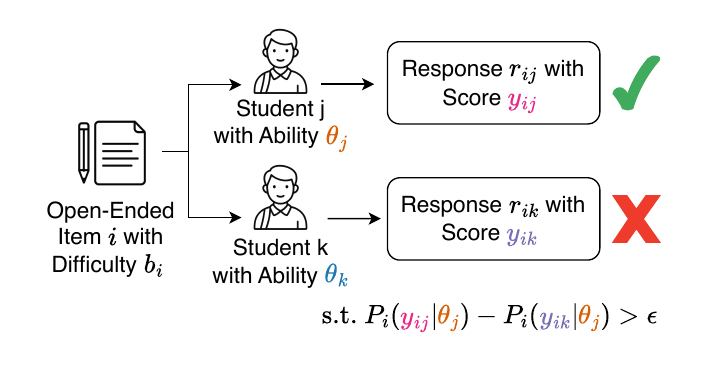}
\caption{Our novel preference pair creation method through comparing response likelihood under IRT.}
\label{fig:dpo}
\end{figure}

\paragraph{Preference Pair Creation.} 
Our novel preference pair creation method is illustrated in Figure~\ref{fig:dpo}. Given an item's question text $q_i$, meta data $m_i$, and a student's ability $\theta_j$, we compare two possible responses: one with a higher IRT likelihood, 
%\nsf{we didn't run my dpo variant using gen responses and scoring model on the new splits, this dpo method uses gt scores}
$r^{w}_{ij}$, which is preferred to another with a lower IRT likelihood $r^{\ell}_{ij}$, thereby forming a preference pair.
We find such preference pairs among the human-scored real student responses. % our large bank of $85$K
For each real student response $r_{ij}$, we construct a set of possible \textit{negative candidates}, $\mathcal{R}_{ij}$, which are other responses to item $i$ that a student with ability $\theta_j$ would be less likely to write, according to the IRT model. Formally:
% \nsf{@al}
% \ml{to make it look nicer, should drop some likelihood expression here in align environment, with the scoring model inside, etc.}
\begin{align}
    \mathcal{R}_{ij} = \left\{ r_{ik} : P_i(y_{ij}|\theta_j) - P_i(y_{ik}|\theta_j) > \epsilon \right\},\label{eq:pref-pairs}
\end{align}
where $P_i$ is either $P^\text{GPCM}_i$ (Eq. \ref{eq:gpcm}) if the scores are ordinal or $P^\text{Bern.}_i$ (Eq. \ref{eq:bernoulli}) if the scores are continuous. The hyperparameter $\epsilon \in [0,1)$ sets the threshold for how much \textit{more} likely a response must be than another to form a preference pair, which is needed to reduce noisy preference pairs. In summary, for some student that responds with $r_{ij}$, for every other possible response, we ask the question: \textit{how likely is it for the same student to write response $r_{ik}$}?
% we first calculate its IRT likelihood $\mathcal{L}^{\text{IRT}}_{ij}$ under its associated student ability $\theta_j$, item difficulty $b_i$, and human-labeled score $y_{ij}$, using the GPCM model in Equation~\ref{eq:gpcm}.
% We then calculate the likelihood $\mathcal{L}^{\text{IRT}}_{ik}$ of every other student response $r_{ik}$ to the same item $i$, but assuming it was written by a student with the same ability $\theta_j$ who wrote $r_{ij}$, thereby asking the question: \textit{how likely is it for the same student to write response $r_{ik}$?}
% We treat $r_{ik}$ as a negative candidate for $r_{ij}$ if the student writing $r_{ij}$ has a lower likelihood under the IRT model $r_{ik}$ by some margin $\epsilon$, i.e., $\mathcal{L}^{\text{IRT}}_{ij} - \mathcal{L}^{\text{IRT}}_{ik} > \epsilon$. 
%\ml{need to add a statement to say which response is the winning one and which is the losing one, otherwise it can be confusing: $^{w/l}_{ij}$ above and $_{ij/k}$ here}
We randomly select $m$ negative candidates from $\mathcal{R}_{ij}$ for training, creating $m$ preference pairs where the ground-truth response, $r^{w}_{ij}=r_{ij}$, is preferred over each sampled response, $r^{\ell}_{ij}=r_{ik}$. We note that $\mathcal{R}_{ij}$ could be empty in some cases, for instance, if the ground-truth student response is unlikely for the corresponding $\theta_j$.
% with the real response $r_{ij}$ being the preferred response $r^{w}_{ij}$ over each negative candidate $r_{ik}$ as the dispreferred response $r^{l}_{ij}$.

% \paragraph{Model Training.}
% SMART further trains the finetuned LLM simulator $S(\cdot)$, described in Section~\ref{subsec:responsegensft}, using the mined preference pairs for DPO training. DPO training helps the simulator $S(\cdot)$ contrast student responses given the same input ability and item, using the following objective:
% \begin{align}
%     \label{eq:dpo}
%     &\mathcal{L}_\text{DPO}(\phi) = -\mathbb{E}_{(x_{ij}, r_{ij}^w, r_{ik}^l) \sim \mathcal{D}} \nonumber\\ 
%         &\left[ \log \sigma 
%         \left( \beta \log \frac{S_\theta(r_{ij}^w|x_{ij})}{S_\text{ref}(r_{ij}^w|x_{ij})} - \beta \log \frac{S_\theta(r_{ik}^l|x_{ij})}{S_\text{ref}(r_{ik}^l|x_{ij})} \right) \right] \nonumber
% \end{align}

% \noindent where $r_{ij}^w$ is the preferred student response with a higher likelihood than the dispreferred response $r_{ik}^l$, and $x_{ij}$ is the input prompt containing the item passage $p_i$, question $q_i$, and student ability $\theta_j$. $S_\theta(\cdot)$ refers to the current student simulator being trained by DPO, $S_\text{ref}(\cdot)$ denotes the SFT-initialized LLM student simulator explained in Section~\ref{subsec:responsegensft}, and $\beta$ is a hyperparameter in DPO training which controls the divergence between learned and reference policies. \ml{standard DPO stuff, can move to appendix if need space}\nsf{\cmark}

% SUBSECTION

\subsection{Scoring Model for Open-ended Responses}

Given a distribution of varied generated student responses to items, we need to score them before fitting an IRT model to estimate item difficulties. 
%Unlike existing work~\cite{feng2025reasoningsamplingaugmentedmcqdifficulty,simstudentsmarter,?} which simulates responses to MCQs and uses a rule-based scoring model to assign scores, we face the additional challenge of training a rubric-based scoring model to assign partial credit to simulated short answer responses to open-ended items. 
%\ml{repeated! no need to state again if we need space}
This scoring task is challenging since it requires scoring synthetic responses to items not seen during training.
We parameterize our scoring model $R(\cdot)$ by finetuning a generative LLM, given its ability to handle long-context inputs, 
%(item passages and rubrics are $X$ and $Y$ tokens on average, respectively), \ml{again a detail that's specific to your data, which we should defer till after the dataset details paragraph} 
to discern complex (partial) scoring criteria, optionally from a rubric, and to generalize to scoring responses to new items in the test set.
We combine the question text $q_i$, the meta data $m_i$, the scoring rubric $w_i$, and the student response text $r_{ij}$ into a prompt $x_{ij}$, and instruct the LLM to only output a single integer for the score label $y_{ij} \in \{0,\ldots,C-1\}$. For ordinal scores, we produce an estimate score $\hat{y}_{ij}$ via greedy decoding from $R(\cdot|x_{ij})$. For continuous scores, following \cite{scarlatos2024exploringknowledgetracingtutorstudent}, we smoothly interpolate between the token logits corresponding to fully correct and incorrect scores:
\begin{align}
    \hat{y}_{ij} = \frac{e^{R(1|x_{ij})}}{e^{R(0|x_{ij})} + e^{R(1|x_{ij})}},
\end{align}
where $R(y|x_{ij})$ is the output logit value for token $y$ returned by $R(\cdot|x_{ij})$, and we train using binary cross-entropy loss. See the exact prompt we use for our scoring model in Appendix~\ref{sec:prompts}.

% SUBSECTION

\subsection{Simulation-Based Difficulty Prediction}
%\nsf{self note: use question or item consistently}

We now detail how to use our ability-controlled student simulator and automated scoring model to predict item difficulties. 
First, we create a population of $n$ simulated students such that their abilities match the training distribution: we create a 50-bucket histogram of abilities in the train set, proportionally draw $n$ samples from the resulting distribution, and convert each sample to a random uniform value in the corresponding bin.
Second, we perform the student simulation: for each drawn ability and item in the test set, we prompt the simulated student model $S(\cdot)$ to generate a corresponding response using nucleus sampling. Third, we score each generated response using the automated scoring model $R(\cdot)$. Finally, we re-fit the IRT model $I(\cdot)$ using the new set of scored responses. 
For each generated response, we assign a student ID by rounding the ability to the nearest decimal place, which we find to be crucial in preventing overfitting by reducing the number of ability parameters significantly. 
We provide additional details on IRT fitting in Appendix~\ref{sec:appendix-irt}.
% In this step, we also include the ground truth student responses from the training set and fit the IRT model on both real and generated responses simultaneously, to keep the IRT model close to its calibrated version fitted on the training set. 
%\al{Honestly I'm not sure this last previous part makes sense, but it does weirdly improve performance} \ml{perhaps!} 
We take the difficulty estimates under this IRT model for the new, unseen items as our final predictions.

% SECTION

\section{Experimental Evaluation}

% SUBSECTION
We now detail experiments that we conduct to validate our approach on real student response datasets.

\subsection{Datasets}

We conduct our experiments on two real-world open-ended student response datasets, each containing around 50 items and thousands of student responses.
% We note that the small number of items constitutes a \textit{low resource setting with few training items}, which is particularly challenging for existing state-of-the-art difficulty prediction methods~\cite{alkhuzaey2024text,benedetto2023quantitative,benedetto2023survey}, which finetune encoder-based pretrained language models (PLMs) that rely on a large number of training items. 

\paragraph{Smarter Balanced.}

We first use a private real-world dataset, provided to us by Smarter Balanced\footnote{\url{https://smarterbalanced.org/}}. The dataset contains student responses to open-ended short answer items administered to students in the United States, containing around $85$K responses to $49$ items from around $63$K students. 
%\nsf{leverage say idea is still generic though shines in this setting, move beginning of meth to here}
%\ml{this stuff belongs till later when you discuss the dataset}
These open-ended items assess reading comprehension skills including argumentation, explanation, and narration, at the grade $6$ level with the average length of student responses being $92$ tokens. Each response is manually scored by a trained human scorer, using a rubric with detailed scoring criteria, with scores in the set $\{0, 1, 2\}$. The meta data for each item includes a passage that the student must read before answering the question. Our trained 3B parameter scoring model achieves a quadratic weighted kappa (QWK) of $0.616$ on this dataset, indicating moderate-to-high agreement with ground-truth scores, which is competitive with unseen item open-ended response scoring on other datasets\footnote{Best QWK on unseen items in NAEP Reading Automated Scoring Challenge was $0.528$: \url{https://github.com/NAEP-AS-Challenge/reading-prediction/blob/main/results.md}}. We show example items in Tables~\ref{tab:qual_ex_sb} and~\ref{tab:qual_ex_sb_2}.

\paragraph{CodeWorkout.}

We additionally use the publicly available CodeWorkout dataset \cite{codeworkout}, originally used for the 2nd CSEDM Data Challenge \cite{csedm}. The dataset, commonly used for student modeling in open-ended programming tasks \cite{liu2022open,duan2025automatedknowledgecomponentgeneration}, contains 50 Java coding problems in a first year college programming course, with student solutions across 2 semesters. Students can attempt a problem multiple times after receiving compiler feedback, so we only consider the first attempt for a problem for each student, resulting in 10,834 responses from 246 students. Each response is scored with a continuous value between 0 and 1, representing the portion of test cases that passed; the test cases themselves are not made available. To ensure that models understand how to correctly solve the problems, we include 2 LLM-generated solutions in each item's meta data, using the method described in \cite{duan2025automatedknowledgecomponentgeneration}. Our trained 3B parameter scoring model achieves a Pearson's correlation coefficient (PCC) of $0.849$ on this dataset, indicating high agreement with ground-truth scores. We show an example item in Table~\ref{tab:qual_ex_codeworkout}.

We split each dataset into $5$ cross-validation folds, with each fold split 60/20/20 item-wise for train/validation/test, respectively. We ensure that difficulties of items across all splits and folds are roughly even. We fit IRT models ($P^\text{GPCM}$ for Smarter Balanced and $P^\text{Bern.}$ for CodeWorkout) on each fold to estimate ground-truth student ability and item parameters.
For each fold, to ensure no information from unseen items is leaked at train time, we fit three separate IRT models, one on train data, one on train plus validation data, and one on all data. 
We use the estimated parameters from the three models to train our student simulator, and to perform validation and test-time evaluation, respectively. We provide further details on IRT fitting in Appendix~\ref{sec:appendix-irt} and dataset processing in Appendix~\ref{sec:appendix-dataset}.

% SUBSECTION

\subsection{Metrics}

\paragraph{Item Difficulty Prediction.}

Following prior work in item difficulty prediction~\cite{tacnn,he2024psychometric}, we use the widely adopted \textit{Pearson Correlation Coefficient} (\textbf{PCC}) to measure the linear correlation between predicted and ground-truth item difficulties, \textit{Spearman Rank Correlation Coefficient} (\textbf{SCC}) to measure the correlation between ranks of predicted and ground-truth item difficulties, and \textit{Root Mean Squared Error} (\textbf{RMSE}) for a precision comparison between predicted and ground-truth item difficulties.
%\nsf{Correlation: PCC, SCC, Kendall tau, QWK, R2 coefficient. Mean: RMSE, MSE, SMD}

\paragraph{Student Simulation.}

We additionally propose metrics to investigate how faithful the generated responses are to real student ones. Since we model students at a population level, reference-based metrics such as ROUGE-L~\cite{lin-2004-rouge} and BERTScore~\cite{bert-score} are not suitable.
% Student responses, from different students even when they have similar abilities, are highly diverse in both style and content, making pairwise textual similarity metrics like ROUGE-L F1~\cite{lin-2004-rouge} and BERTScore F1~\cite{bert-score} unsuitable to evaluate the student simulator. 
Therefore, following prior work in population simulation~\cite{bui2025mixtureofpersonaslanguagemodelspopulation,yu2023diversity}, we report distribution-level metrics comparing the generated and ground-truth response distributions in a latent embedding space.
We first extract embeddings for all responses using a sentence encoder~\cite{reimers-2019-sentence-bert}, specifically \texttt{multi-qa-mpnet-base-dot-v1}.
We then compute the \textit{Fréchet Inception Distance} (\textbf{FID})~\cite{heusel2017gans}, which is common for evaluating generated images but has recently been adapted to text \cite{yue-etal-2023-synthetic,bui2025mixtureofpersonaslanguagemodelspopulation}, and \textbf{MAUVE}~\cite{pillutla2021mauve}, which measures the similarity of cluster distributions in the embedding space, to evaluate the alignment between the predicted and ground-truth student response distributions. 
Following~\cite{bui2025mixtureofpersonaslanguagemodelspopulation}, we also measure the similarity in diversity distribution by computing \textbf{Div. KL}, the KL divergence between the histograms of pairwise response cosine similarity across all ground-truth responses and across all simulated responses.
%\al{honesty I hate the name they pick for this metric since it's really non-descriptive. Can we call it Diversity KL instead?} 
%\ml{/giphy puke}
%\nsf{yeah it's not intuitive. however, lower is better for this metric and diversity KL somehow sounds like higher the better, since this checks diversity alignment, but yeah pick the one you like}
% Diversity KL first computes the cosine similarities between all pairs of predicted responses, and ground-truth responses, and bins them into two separate histograms. The predicted and ground-truth all-pair cosine similarity histograms are then normalized to a probability distribution followed by computing the KL divergence between them. \al{can probably put details in appendix}
Finally, we compute \textbf{$\theta$-Align}, the Spearman correlation between input student ability and the predicted score of the corresponding generated response, measuring how well simulated students align with their prompted abilities.

% SUBSECTION

\subsection{Baselines}

We compare our approach, SMART, to state-of-the-art item difficulty prediction approaches as well as to several strong novel baselines adapted to our setting. These approaches fall into two categories: predicting difficulties 1) directly from item text, and 2) via student simulation.

\paragraph{Direct Difficulty Prediction Baselines.}

Following prior work~\cite{alkhuzaey2024text,benedetto2023quantitative,benedetto2023survey,kapoor2025predictionitemdifficultyreading,north2025cyborgdatamerginghuman},
%\ml{does chris have a preprint paper on this? using modernbert for scoring?} \al{don't think it's released yet} 
we adopt a strong finetuning baseline, \textbf{Diff SFT}, which finetunes a recent and capable encoder-based language model, specifically ModernBERT-base~\cite{modernbert}. We use a regression setup, minimizing mean squared error loss, to directly predict item difficulties given the item's question text and meta data. We also adopt a \textbf{Random Forest} baseline, which is common for difficulty prediction \cite{alkhuzaey2024text,gombert-etal-2024-predicting}, where we encode the item's question text and meta data using the \texttt{multi-qa-mpnet-base-dot-v1} SBERT model~\cite{reimers-2019-sentence-bert}, and use the resulting embeddings as input features to fit a random forest regressor to item difficulty.
%\ml{what's the objective for this SFT? you minimize what?} 

Since our set of items is small, we also introduce several baselines that do not require training. We use a simple yet strong baseline, \textbf{kNN Mean}, that retrieves the $k$ items in the train set with the highest SBERT embedding cosine similarity, and takes their mean difficulty as the predicted test item difficulty. Similar methods have been used in automated scoring~\cite{bin2008automated,gomaa2012short}. We search over $k \in [1,5]$ and select the best performing value on the validation set, selecting $k=1$ for Smarter Balanced and $k=5$ for CodeWorkout.
Finally, we adopt two LLM prompting baselines leveraging in-context learning (ICL)~\cite{gpt3icl}, \textbf{ICL Random} and \textbf{ICL kNN}, where we prompt \texttt{Llama-3.1-70B-Instruct} with $k$ in-context examples of items retrieved from the train set at random or using k-Nearest Neighbors (kNN)~\cite{feng-etal-2024-exploring}, respectively. We ask the model to first reason about the question's difficulty in a chain-of-thought manner before generating a numeric difficulty prediction. For kNN retrieval, we use the same SBERT model as above, and perform a similar hyperparameter search, selecting $k=4$ for Smarter Balanced and $k=2$ for CodeWorkout.
% Each item example contains its associated passage, question, and numeric item difficulty.
% For kNN retrieval, we embed the test passage and question text using SBERT~\cite{reimers-2019-sentence-bert}, specifically \texttt{multi-qa-mpnet-base-dot-v1}, and retrieve the $k$ nearest training items in the same embedding space. We experiment with $k \in [1,5]$ on the validation set and find that $k=4$ works best. We prompt the model to first reason about the question's difficulty in a chain-of-thought manner before generating its numeric prediction.
% We also compare against \textbf{SBERT kNN Mean}, a simple yet strong baseline, which retrieves the closest $k$ training items in similarity to the target test item using kNN retrieval as above, and uses their average difficulty as the predicted test item difficulty \cite{?}. We experiment with $k \in [1,5]$ on the validation set and find that $k=1$ works best, i.e., simply copying the difficulty of the most similar item. Finally, we compare against \textbf{SBERT RF}, where we train a random forest regressor to predict item difficulty using the SBERT features as input. Random forest methods have been shown to be highly effective in difficulty prediction~\cite{?}.

\paragraph{Student Simulation-based Difficulty Prediction Baselines.}

We compare SMART to other student simulation-based approaches for item difficulty prediction. For a fair comparison, we use the same base model and prompt template as SMART for the following methods, allowing these baselines to also serve as ablations.
First, we employ \textbf{Response Gen ZS}, where we prompt an LLM in a zero-shot (ZS) manner to generate a student response given the item and ability. Since the model has not been trained on the distribution of numeric student abilities, we convert abilities to corresponding labels from 10 descriptive buckets shown in Appendix~\ref{sec:appendix-baselines}, ranging from ``minimal'' to ``mastery'', with descriptions obtained by prompting Claude 3.5 Sonnet~\cite{claude-3.5-sonnet}. This method is similar to the persona-based prompting method in~\cite{he2024psychometric}, where we use ability instead of student demographics.
%\al{should probably put all buckets/ranges in the appendix. also we should ask Chris where he got these (hopefully we can cite something)}
Second, we employ \textbf{Response Gen SFT}, as explained in Section~\ref{subsec:responsegensft}, which finetunes the LLM on ground-truth student responses given the item and ability. Prior works that finetune simulated students generally include prior student responses in the prompt~\cite{he2024psychometric,zelikman-etal-2023-generating}; we use the student's ability as a proxy for this information.
%\nsf{DPO-$\theta$?} 
%\al{to address the fact these all sound like ablations, we could mention that no prior work has done trainable student simulation for open-ended responses. some have done simulation with varying scales of pre-trained models, but this is pretty similar to our zero-shot setting (difference is we're using one model but with different prompts)}

\begin{table*}[th]
\small
\centering
\begin{tabular}{p{0.24\linewidth}C{0.08\linewidth}C{0.08\linewidth}C{0.08\linewidth}|C{0.08\linewidth}C{0.08\linewidth}C{0.08\linewidth}}

\toprule

\multirow{3}{*}{Model} & \multicolumn{3}{c}{Smarter Balanced} & \multicolumn{3}{c}{CodeWorkout}\\
\cmidrule{2-7}
& PCC $\uparrow$ & SCC $\uparrow$ & RMSE $\downarrow$ & PCC $\uparrow$ & SCC $\uparrow$ & RMSE $\downarrow$\\

\midrule

\rowcolor{gray!21} \multicolumn{7}{c}{Direct Item Difficulty Prediction Methods}\\
Diff SFT (ModernBERT) & $0.4937$ & $0.3188$ & $0.7627$ & $-0.0425$ & $-0.0036$ & $0.7714$\\ 
Random Forest (SBERT) & $0.4021$ & $0.3115$ & $0.8623$ & $0.1846$ & $0.1418$ & $0.6783$\\ 
kNN Mean (SBERT) & $0.5959$ & $0.5219$ & $0.6806$ & $0.1949$ & $0.2415$ & $0.6701$\\ 
ICL Random (Llama-3.1-70B) & $0.1215$ & $0.0348$ & $1.0426$ & $0.2487$ & $0.1806$ & $0.6462$\\ 
ICL kNN (Llama-3.1-70B) & $0.5273$ & $0.4717$ & $0.7634$ & $0.0209$ & $0.0821$ & $0.7397$\\ 

\rowcolor{gray!21} \multicolumn{7}{c}{Student Simulation-based Item Difficulty Prediction Methods (Llama-3.2-1B)}\\
Response Gen ZS & $0.2651$ & $0.2436$ & $0.9088$ & $0.0733$ & $0.0812$ & $0.7226$\\
Response Gen SFT & $0.5621$ & $0.4254$ & $0.7174$ & $0.2475$ & $0.2388$ & $0.6487$\\
SMART (ours) & $\underline{0.6490}$ & $0.4861$ & $\underline{0.6699}$ & $0.3082$ & $0.3188$ & $0.6225$\\

\rowcolor{gray!21} \multicolumn{7}{c}{Student Simulation-based Item Difficulty Prediction Methods (Llama-3.2-3B)}\\
Response Gen ZS & $0.3446$ & $0.3963$ & $0.8568$ & $0.1487$ & $0.1636$ & $0.6912$\\
Response Gen SFT & $0.6069$ & $\textbf{0.5830}$ & $0.6905$ & $\textbf{0.4256}$ & $\underline{0.3648}$ & $\textbf{0.5627}$\\
SMART (ours) & $\textbf{0.6737}$ & $\underline{0.5661}$ & $\textbf{0.6197}$ & $\underline{0.3929}$ & $\textbf{0.4230}$ & $\underline{0.5890}$\\
\bottomrule
\end{tabular}
%\vspace{-.2cm}
\caption{Performance on question difficulty prediction. Best performance is in \textbf{bold} and second best is \underline{underlined}. SMART outperforms all direct prediction methods and generally outperforms other simulation-based methods.}
\label{tab:results-diff}
%\vspace{-.2cm}
\end{table*}

% SUBSECTION

\subsection{Experimental Setup}

We use \texttt{Llama-3.2-Instruct}~\cite{grattafiori2024llama} from HuggingFace~\cite{wolf2020huggingfacestransformersstateoftheartnatural} as the base model for our simulated student and automated scoring models. We perform all experiments with both the 1B and 3B parameter versions to examine how our methods scale across model sizes. We use the 1B automated scoring model to score responses from the 1B simulated student model, and do the same for 3B. We load base models with NF4 quantization~\cite{dettmers2023qlora} and finetune using LoRA~\cite{hu2022lora}. We provide further experimental details, including all hyperparameters, in Appendix~\ref{sec:smart-hyperparams}.

\section{Results, Analysis, and Discussion}

We now detail our quantitative results on item difficulty prediction and student response simulation, and qualitatively analyze the simulated student responses and SMART's strengths and weaknesses.

% SUBSECTION

\subsection{Quantitative Results}

\paragraph{SMART outperforms other methods on item difficulty prediction.}
We report the average performance across all test items and cross-validation folds for both datasets in Table~\ref{tab:results-diff}. Our approach, SMART, outperforms all direct prediction methods on all metrics, even largely maintaining this advantage when using the small 1B base LLM. It also outperforms all other student simulation methods, with the exception of SFT when using 3B models; in these cases, SMART outperforms SFT by wide margins on some metrics while it is closely behind on others. These results indicate that both methods are powerful predictors of item difficulty, although there is high variability due to the small number of items in each dataset; larger test sets may be required to more clearly show SMART's advantage. We also observe that CodeWorkout is a more challenging dataset, with most methods performing worse than they do on Smarter Balance in terms of PCC and SCC. Still, the fact that SMART consistently outperforms baselines on both datasets shows that it can generalize across educational domains and student populations.

\begin{table*}[th]
\small
\centering
\begin{tabular}{p{0.15\linewidth}C{0.085\linewidth}C{0.06\linewidth}C{0.08\linewidth}C{0.08\linewidth}|C{0.085\linewidth}C{0.06\linewidth}C{0.08\linewidth}C{0.08\linewidth}}

\toprule

\multirow{3}{*}{Model} & \multicolumn{4}{c}{Smarter Balanced} & \multicolumn{4}{c}{CodeWorkout}\\
\cmidrule{2-9}
& MAUVE $\uparrow$ & FID $\downarrow$ & Div. KL $\downarrow$ & $\theta$-Align $\uparrow$ & MAUVE $\uparrow$ & FID $\downarrow$ & Div. KL $\downarrow$ & $\theta$-Align $\uparrow$\\

\midrule

\rowcolor{gray!21} \multicolumn{9}{c}{Student Simulation-based Item Difficulty Prediction Methods (Llama-3.2-1B)}\\
Response Gen ZS & $0.0058$ & $0.2713$ & $0.2535$ & $0.0399$ & $0.0196$ & $0.0747$ & $0.0714$ & $-0.0014$\\
Response Gen SFT & $0.1483$ & $0.0847$ & $\underline{0.0215}$ & $0.4481$ & $\underline{0.1777}$ & $\underline{0.0269}$ & $\underline{0.0315}$ & $0.1730$\\
SMART (ours) & $0.1315$ & $0.0918$ & $0.0482$ & $\underline{0.7338}$ & $\textbf{0.1931}$ & $\textbf{0.0267}$ & $\textbf{0.0284}$ & $\textbf{0.5895}$\\

\rowcolor{gray!21} \multicolumn{9}{c}{Student Simulation-based Item Difficulty Prediction Methods (Llama-3.2-3B)}\\
Response Gen ZS & $0.0052$ & $0.3386$ & $0.4032$ & $0.0211$ & $0.0042$ & $0.1284$ & $0.2779$ & $-0.0101$\\
Response Gen SFT & $\textbf{0.1858}$ & $\textbf{0.0757}$ & $\textbf{0.0137}$ & $0.5158$ & $0.1426$ & $0.0333$ & $0.0515$ & $0.2226$\\
SMART (ours) & $\underline{0.1713}$ & $\underline{0.0785}$ & $0.0279$ & $\textbf{0.7655}$ & $0.1479$ & $0.0303$ & $0.0335$ & $\underline{0.5582}$\\
\bottomrule
\end{tabular}
%\vspace{-.2cm}
\caption{Performance on student simulation metrics. Best performance is in \textbf{bold} and second best is \underline{underlined}. SMART significantly outperforms other methods on ability following.}
\label{tab:results-sim}
%\vspace{-.2cm}
\end{table*}

\paragraph{Student simulation is key when working with few training items.}
We observe that direct prediction baselines significantly underperform student simulation methods on both datasets. Diff SFT, which follows the approach of existing state-of-the-art methods~\cite{alkhuzaey2024text,benedetto2023quantitative,benedetto2023survey}, performs moderately well on Smarter Balanced but poorly on CodeWorkout. This result is perhaps not surprising, since Diff SFT's training signal completely comes from given item difficulties, and thus can only work when a large number of training items are available. Direct difficulty prediction baselines that require no training also follow a similar pattern: kNN Mean and ICL kNN perform moderately on Smarter Balanced, likely because reading comprehension questions with similar semantic features are likely to pose similar tasks to the student and thus have similar difficulties. However, since this heuristic does not apply to coding questions, these baselines do not generalize across datasets. The poor performance of ICL Random also shows that powerful LLMs do not perform well on the difficulty prediction task without careful calibration. In our low resource setting with few items to train with, student simulation is key: student ability and item difficulty are only related via the response score when training an IRT model, but our student simulator can effectively couple IRT training with textual student responses and a scoring model. As a result, this extra information helps us to calibrate the IRT model better, making it easier to generalize to previously unseen items. 

%to generate synthetic simulated responses at scale on new items in the test set, which are then IRT fitted for accurate item difficulty prediction.  \nsf{fix intuition}.

\paragraph{Ablation study.}
We compare the performance of Response Gen ZS, Response Gen SFT, and SMART, with all models parameterized with the same base LLM and prompts, serving as an ablation study. Response Gen SFT significantly outperforms Response Gen ZS, which suggests that finetuning is required for the LLM to adapt to the response distribution of real student responses. With the exception of some metrics when using 3B models, SMART outperforms Response Gen SFT, which validates our hypothesis that preferring responses with high IRT likelihoods in DPO training helps in the difficulty prediction task by calibrating the model in the context of student ability and implicit item difficulty.
% the student simulator to better align with the desired student ability as input control. %In terms of model scaling trends, within the \texttt{Llama-3.2}~\cite{grattafiori2024llama} family, methods parameterized with the bigger $3$B parameter base LLM performs better than using the $1$B parameter base LLM across all metrics. 

\paragraph{SMART simulated student responses are aligned with abilities and diverse.}
%Table~\ref{tab:results-student-sim} 
Table~\ref{tab:results-sim} compares the student simulation methods on metrics regarding the quality and faithfulness of simulated responses. We see that SMART outperforms other methods by a wide margin on the $\theta$-Align metric, which measures the correlation between the desired input ability and inferred ability by scoring the generated responses. This observation shows that DPO training leads to a strong preference for the model to follow the input ability, resulting in a faithful student simulator. 
We also compare the simulated student methods on distribution similarity metrics, measuring how closely their generated responses align with ground-truth ones. 
Response Gen ZS performs poorly, which emphasizes the need for finetuning or DPO training, over simple prompting, to align the LLM with diverse responses by real students. 
Interestingly, Response Gen SFT is marginally better than SMART on these metrics on Smarter Balance, while SMART is better on CodeWorkout. This observation highlights the fact that DPO training prioritizes the IRT likelihoods, which may or may not align with how realistic the generated responses appear. We also observe that 1B models outperform 3B models on these metrics on CodeWorkout, perhaps due to the lower coding ability of the 1B model, therefore aligning better with code written by novice coders (students).

\subsection{Qualitative Evaluation}

To further understand the strengths and weaknesses of our method, we perform a qualitative analysis, where we examine patterns in items and generated responses that affect difficulty prediction accuracy.

\paragraph{Models Reflect Ability-Specific Features in Student Responses.}
%The main advantage of our method is the ability to leverage real student responses and their likelihoods under IRT, to learn how different student ability levels are reflected in responses to different items. We see that many ability-specific linguistic features are reflected in the model-generated responses. 
%indicating that the model has picked up on key patterns in the data that result in score distributions that are highly aligned with the ground-truth.
Tables~\ref{tab:qual_ex_sb} and~\ref{tab:qual_ex_sb_2} show examples of generated responses to Smarter Balanced items for students with varying ability levels. Compared to SFT, SMART more clearly reflects realistic patterns based on ability.
Broadly speaking, students with low ability have short one-sentence responses, spelling mistakes, poor grammar, and incorrect punctuation. 
High-ability students, on the other hand, have longer responses with complex sentence structures and strong vocabulary, and make an effort to answer all aspects of the question to receive full credit.
SMART also captures student response diversity, especially for medium-ability students, whose responses usually contain simple words, but differ in length, coherence, and score obtained by the trained scoring model. 
Our preference pair creation encourages this diversity since varied responses can be preferred as long as they are sufficiently likely under the IRT model.

Table~\ref{tab:qual_ex_codeworkout} shows examples of real and generated responses to a CodeWorkout item. SMART's difficulty prediction is much more accurate than SFT's on this item, primarily due to the diversity of generated responses; the SFT responses are semantically similar and receive similar scores across ability levels, while for SMART, low ability students score poorly while medium and high ability students score highly. Additionally, SMART more faithfully reflects semantic patterns in the student code, such as medium ability responses being longer and high ability responses being more concise and using more clear coding constructs.

\paragraph{Performance is Item-dependent.}
For Smarter Balanced items, we find that SMART's performance is highly dependent on the type of reading comprehension task (argumentation, narration, explanation, among others) in the item. 
In particular, difficulty estimates are accurate for items that require information recall or those that add narrative elements to a story. 
On the other hand, estimates are less accurate for items requiring higher order thinking~\cite{lewis1993defining}, such as summarizing the main idea of a passage, or writing an introduction to a paragraph that includes a ``thesis'' or a ``main controlling idea''. 
For these items, we find that SMART tends to \textit{overestimate} their difficulty. We see that even for high input ability values, SMART tends to generate incomplete responses that only state facts from the passage, rather than summarizing it. Such responses receive only partial credit and thus lead to high difficulty values under IRT. Future work can attempt to augment the training data with synthetic responses with higher-order thinking to aid simulation for high-ability students. 
%generated responses from high-ability simulated students are incomplete with a focus on stating facts from the passage rather than coherently summarizing themes as required by the item, and thus receive only partial credit. These response patterns for higher-order thinking-type questions could be due to inherent limitations in the base LLM. Future work could explore larger base LLMs and also cater negative example selection to countering this behavior when forming preference pairs.
%overfitting to responses that emphasize information recall, or could be due to limitations in the base model with being able to draw effective conclusions from information. 
%Future work could attempt to solve this issue by including generated responses in the training set, forming negative samples from model-generated high ability student solutions that do not sufficiently solve the given task.

\paragraph{Outlier Generated Responses.}
We observe that SMART occasionally generates outlier responses with fidelity issues, with failure patterns including 1) repeating the passage, 2) not being relevant to the item, and 3) being highly repetitive.
These outlier responses are often inaccurately scored by the scoring model, affecting downstream item difficulty prediction. 
A possible solution is to use a fidelity checking model to filter out such outlier responses and re-generate when it happens. 

\section{Related Work}

%\nsf{move to apdx, reorganize to two subsections: diff pred and student sim}
% SUBSECTION

\paragraph{Direct Item Difficulty Prediction}
%\ml{move to last!}
% Commonly done for MCQs \cite{yaneva-etal-2024-findings,feng2025reasoningsamplingaugmentedmcqdifficulty}. Feature-based \cite{tack-etal-2024-itec}. BERT-based \cite{gombert-etal-2024-predicting,zhou2020multi,loginova-etal-2021-towards,benedetto-etal-2021-application}. TACNN \cite{huang2017question}. LLM student simulation to mine for features \cite{duenas-etal-2024-upn}. Train group of LSTM's on varying proportions of incorrect labels to get simulated students of varying ability for difficulty prediction \cite{lalor-etal-2019-learning}. Prompt a large set of varying size/strength pre-trained (L)LMs and fit IRT model to get difficulty estimates \cite{liu-leveraging-2025,uto2024question}. Train simulated students by conditioning on previous responses by that student and use predictions to estimate difficulty and response time for true/false items \cite{zelikman-etal-2023-generating}. Closest to our work is \cite{he2024psychometric} - they do the simulation pipeline for difficulty estimation using 1) an ensemble of LLMs, 2) demographic-based persona prompting, and 3) fine-tuning; they find that (2) and (3) can be better depending on the dataset; our novelty from them is both in open-ended resps and ability alignment. Our system 1) models open-ended responses (not done in other fine-tuned systems), 2) trains the simulated students to follow a prompted ability explicitly, 3) deals with small item sets (feature-based or BERT-based can't do this well).

Item difficulty prediction has been widely studied \cite{alkhuzaey2024text}, particularly in the context of multiple-choice questions (MCQs) \cite{yaneva-etal-2024-findings,feng2025reasoningsamplingaugmentedmcqdifficulty,reyes2023multiple}. It is common to train regression models using handcrafted features \cite{byrd-srivastava-2022-predicting,tack-etal-2024-itec,kapoor2025prediction,gombert-etal-2024-predicting}, although finetuning language models such as BERT has proven to be highly effective \cite{gombert-etal-2024-predicting,zhou2020multi,loginova-etal-2021-towards,benedetto-etal-2021-application}. Custom architectures, such as TACNN, have also been proposed for this task \cite{huang2017question}.

\paragraph{Simulation-Based Difficulty Prediction}

An increasing body of research uses simulated students to estimate item difficulty.
% For example, \cite{duenas-etal-2024-upn} uses LLMs to generate synthetic responses and mines useful features.
They follow a similar pipeline to our work, where they overgenerate a large set of responses using AI models, and then fit an IRT model to these responses to recover the item difficulty. Such methods have prompted ensembles of pre-trained language models to answer the items \cite{liu-leveraging-2025,uto2024question}, finetuned language models to predict student responses given prior responses \cite{zelikman-etal-2023-generating}, and trained LSTMs with varying proportions of incorrect labels \cite{lalor-etal-2019-learning}. \citet{he2024psychometric} experiment with LLM ensembles, finetuning, and persona-based prompting, finding that the most effective method depends on the dataset.

Our approach builds on this line of work but introduces several key innovations. First, we explicitly train simulated students to follow a prompted ability level, allowing the simulated students to align with the underlying IRT distribution. Second, we focus on open-ended responses, which are not handled by most finetuned or feature-based methods.  Third, our system is designed to operate effectively on small item sets, where feature-based and BERT-based methods typically underperform.

% SUBSECTION

% \subsection{Simulated Students}

% General LLM-based, mostly in dialogues (MathDial, LearnLM, Shashank papers). Knowledge tracing (classic) and open-ended (OKT/TICTOK, Dialogue KT). Challenges - following ability/knowledge. Our system 1) models open-ended responses, 2) trains simulated students to follow given ability.
% \begin{enumerate}
%     \item Generative Students: Using LLM-Simulated Student Profiles to Support Question Item Evaluation - https://arxiv.org/abs/2405.11591
%     \item Psychometric Alignment: Capturing Human Knowledge Distributions via Language Models - https://arxiv.org/abs/2407.15645
%     \item Leveraging LLM respondents for item evaluation: A psychometric analysis - \url{https://bera-journals.onlinelibrary.wiley.com/doi/10.1111/bjet.13570?af=R}
%     \item Generating and Evaluating Tests for K-12 Students with Language Model Simulations: A Case Study on Sentence Reading Efficiency - https://arxiv.org/abs/2310.06837
%     \item \url{https://www.frontiersin.org/journals/education/articles/10.3389/feduc.2024.1494431/full}
%     \item Field-Testing Multiple-Choice Questions With AI Examinees: English Grammar Items
% \end{enumerate}

% SECTION

\section{Conclusions and Future Work}

In this paper, we presented SMART, a novel method for aligning LLM-based simulated students with a prompted ability and implicit item difficulty via DPO training. The key is to form preference pairs where preferred responses are more likely under an IRT model. We perform a large-scale simulation with the resulting student model, automatically scoring generated responses with a trained LLM, and fit the resulting data with IRT to estimate the difficulties of unseen items.
% The key is to use DPO to align the simulated responses by SMART with student ability, by creating preference pairs that prefer responses with a high likelihood under the IRT model.
Through extensive experiments on a real-world student response dataset, we show that SMART outperforms other item difficulty prediction methods by capturing response patterns in the data that reflect the diverse abilities of the students and the difficulties of the items.

There are many avenues for future work. First, we can include generated student responses
% from the SFT-initialized student simulator, Response Gen SFT,
in the pool of candidates for DPO~\cite{parikh2025lookalikeconsistentdistractorgeneration}, which may help to explicitly disprefer outlier responses that are not aligned with real student data. Second, we can explore the applicability of SMART in other domains such as programming, math, and in non question-response settings such as dialogues~\cite{scarlatos2025trainingllmbasedtutorsimprove}. Finally, we can extend SMART's student representation to go beyond simple scalar-valued student abilities and into more descriptive personas~\cite{wu-etal-2025-embracing}. 
%Third, we can extend SMART to other item types including essays with longer student responses to open-ended items.

%A natural extension of our idea is to use generated student responses from the SFT-initialized LLM student simulator to form preference pairs. Preliminary experiments using generated student responses in preference pairs did not improve performance possibly due to the noise introduced by using a learned scoring model, and not human raters, to assign scores. We leave this extension as a promising direction for future work. 

% SECTION
\section*{Acknowledgments}
We thank Smarter Balanced for providing the private dataset used in our experiments. We also thank Joseph Di Garbo, Frank Rijmen, Suhwa Han, Benjamin Godek, Bokhee Yoon, Zhangqi Duan, and Jacqueline Scarlatos for helpful discussions around this work. This work is partially funded by the NSF under grants 2237676 and 2418657. 

% SECTION

\clearpage
\section*{Limitations}

There are several practical limitations to our work. First, while we experiment on two datasets across educational domains and student populations, there are limitations to these datasets. The Smarter Balanced dataset is not publicly available, and the CodeWorkout data takes place over the course of two semesters, making static ability modeling challenging. Additionally, the small number of items in each dataset makes reliable evaluation challenging. To the best of our knowledge, there are no other suitable publicly available open-ended testing datasets. Second, we restrict our student simulation to models no larger than $3$B parameters, which is relatively small. While our methods are effective at this size, it would be helpful to see if scaling to larger sizes is worth the cost. Third, we find that the textual distribution of our generated responses does not always match the ground-truth responses, and that generated responses are occasionally degenerate. As a result, the pool of simulated responses may not look realistic from a human perspective. This limitation restricts our method's usefulness as a more generic student simulator, although it could be addressed by improvements to the method in future work.

% SECTION

\section*{Ethical Considerations}

There are several potential societal benefits to our work. Primarily, accurate automated difficulty prediction can greatly benefit educational assessment: it can reduce the cost of pre-testing to calibrate item difficulties, it can reduce the risk of sensitive items being leaked from pre-tests, and it can enable calibrating difficulty for AI-generated personalized items, enabling more accessibility in assessments. There are several potential risks to our work as well. Incorrectly calibrated items could lead to unfair educational assessments if the calibrations are overfitted to a specific demographic. As bias is common in AI systems, simulated students may not sufficiently represent minority or non-English speaking students, thus leading to calibration errors for these populations. Future work should study these effects before deploying any such simulation-based difficulty prediction system.


\bibliography{custom}

\appendix

% SECTION

\section{Additional Experimental Details}

% SUBSECTION

\subsection{Student Simulation Hyperparameters}
\label{sec:smart-hyperparams}

We perform a preliminary hyperparameter search on our validation sets. We set LoRA's rank = 128, $\alpha$ = 64, and dropout = 0.05. For Smarter Balanced, for both the scoring model and student model, we perform SFT for 1 epoch, with learning rate = 1e-4 and linear warmup for 10\% of steps, weight decay = 1e-2, gradient clipping norm = 1.0, and effective batch size = 64 using gradient accumulation. For DPO, we set learning rate = 1e-5, $\beta$ = 0.5, the probability threshold $\epsilon=0.1$ (Eq.~\ref{eq:pref-pairs}), the number of negative candidates $m=3$, and train on a random 20\% subset of the training set to reduce costs. For CodeWorkout, we use the same hyperparamters except for the following: for the scoring model we perform SFT for 5 epochs with learning rate = 5e-5, for the student model we perform SFT for 3 epochs with learning rate = 1e-5, and perform DPO for 1 epoch with learning rate = 1e-6 on the full training set. We conduct all experiments on the Smarter Balanced dataset on NVIDIA A10G GPUs and all experiments on the CodeWorkout dataset on NVIDIA A40 GPUs.

At inference time, we sample $n=1,000$ student abilities for simulation and generate a single response for each item/ability pair. We decode up to $500$ tokens with nucleus sampling, setting temperature = $0.7$ and p = $0.95$. We use an inference-time batch size of $64$ for the simulated student model and $2$ for the scoring model, optimizing for inference speed.

On the Smarter Balanced dataset, for both SFT and DPO, training the simulated student model takes approximately 3 hours on a single GPU for the 1B model and 6 hours on 2 GPUs for the 3B model. Training the automated scoring model takes approximately 4 hours on a single GPU for the 1B model and 9 hours on 2 GPUs for the 3B model. Generating 10,000 responses (1,000 abilities over 10 items) takes approximately 1 hour on a single GPU for the 1B model and 3 hours on a single GPU for the 3B model. Scoring these responses takes approximately 15 minutes on a single GPU for the 1B model and 35 minutes on a single GPU for the 3B model.

\subsection{Additional Implementation Details}
\label{sec:appendix-baselines}

We use \texttt{multi-qa-mpnet-base-dot-v1} for our SBERT model since it is the highest performing model with a 512 token limit, according to the sentence transformers model page\footnote{\url{https://sbert.net/docs/sentence_transformer/pretrained_models.html\#original-models}}. We also found that it worked better than \texttt{all-distilroberta-v1} in preliminary experiments.

We train our random forest baseline using the RandomForestRegressor class from the sklearn\footnote{\url{https://scikit-learn.org/}} library. We additionally restrict max\_features to the size of train set to avoid overfitting, which we found improves performance.

For our Response Gen ZS student simulation baseline, we first normalize all train abilities to $[-3,3]$ before sampling, and then map abilities to 10 labeled buckets with the following ranges:
``Minimal'': $\theta<-2.5$, ``Emerging'': $\theta \in [-2.5,-2)$, ``Developing'': $\theta \in [-2,-1.5)$, ``Approaching Proficiency'': $\theta \in [-1.5,-0.8)$, ``Proficient'': $\theta \in [-0.8,0)$, ``Competent'': $\theta \in [0,0.8)$, ``Skilled'': $\theta \in [0.8,1.5)$, ``Advanced'': $\theta \in [1.5,2.0)$, ``Exceptional'': $\theta \in [2,2.5)$, ``Mastery'': $\theta\ge2.5$.

For computing MAUVE, we use the official implementation\footnote{\url{https://github.com/krishnap25/mauve}}. For computing FID, we adapt the implementation from~\cite{fid-impl}.

\subsection{Predicted Difficulty Normalization}

For all methods, including baselines, we transform predicted difficulties to have the same mean and standard deviation as the ground-truth difficulties on the train set; doing so improves RMSE when predicted difficulties are correlated but distributionally shifted from the ground-truth. We perform the transformation using the following formula:
\begin{align}
    \hat{b}_i \leftarrow \frac{\sigma_2}{\sigma_1}(\hat{b}_i - \mu_1) + \mu_2,
\end{align}
where $\hat{b}_i$ is the predicted difficulty for item $i$, $\mu_1$ and $\sigma_1$ are the mean and standard deviation of the predicted item difficulties, respectively, and $\mu_2$ and $\sigma_2$ are the mean and standard deviation of the item difficulties in the train set, respectively.

\subsection{IRT Model Fitting}
\label{sec:appendix-irt}

We perform IRT fitting at 2 distinct stages in our pipeline: 1) when fitting to real student responses to create the ground-truth ability and item parameters, and 2) when fitting to simulated student responses to produce difficulty estimates. 

The Smarter Balanced dataset does not contain student identifiers, but rather contains prior ability estimates based on each student's responses to a larger set of questions, including other types such as multiple choice questions; the 49 short answer questions we use comprise all short answer items in the full set. To ensure that other item types and test set responses do not influence our ability estimates, we fit an IRT model from scratch only on the 49 items used in this study. Since we do not have access to student identifiers, we use the prior ability as a proxy. However, the student responses are very sparse on our set of questions, with around $63$K unique ability values for around $85$K responses, leading to a poor fit for the IRT model. As a solution, we round each prior ability to a single decimal place (we found this precision to work best out of several tried) and use the resulting value as the student identifier. This transformation leads to a significantly better fit for the ground-truth IRT model. We note that this process assumes access to prior estimates of student ability, which may or may not be available in real-world settings. However, we note that this a limitation of the data, rather than our method. Additionally, we note that this transformation does not unfairly benefit our method over the baselines, so the comparison between methods in this work remains fair.

When we fit the IRT model with the simulated responses, we perform a similar operation for assigning identifiers. We round the prompted ability for each simulated response to one decimal place, and use this value to form a new identifier for all simulated responses that result in that rounded ability. When re-training the IRT model, we include the ground-truth student responses from the train set only. While these ground-truth student responses do not directly interact with the estimated parameters for non-train items or simulated students, we find their inclusion to improve performance, perhaps due to their effect on global variables such as AdamW's adaptive learning rate.

We implement our IRT models using PyTorch, using the formulations in Equations~\ref{eq:gpcm} and~\ref{eq:bernoulli} to compute the probability of a student's score for a particular item.
% We fix all discrimination parameters to 1, which we found improves model fit. We additionally enforce that the first step parameter equal 0, i.e. $d_{i0} = 0$. Additionally, to ensure that all step parameters sum to 0, we only train a single step parameter for each item, $d_{i1}$, and set the third step to be the negative of the second, i.e., $d_{i2} = - d_{i1}$.
For GPCM, following prior work~\cite{muraki1993information}, we set the first step parameter to $d_{i0}=0$, since only $C-1$ step parameters are required for $C$ score classes. We also impose the location constraint~\cite{muraki1993information}, where all step parameters for an item sum to $0$, i.e., $\sum_{y=0}^{C-1} d_{iy} = 0$, ensuring that difficulty parameters across items are on the same scale and are identifiable. We also fix all discrimination parameters to $1$, which improves model fit. We do not enforce normal priors on any parameters. We train the model using gradient descent, using the cross-entropy loss between the predicted score probabilities and the ground-truth scores. We train using the AdamW optimizer for 50 epochs with learning rate = 1e-3, weight decay = 0, and batch size = 256; for CodeWorkout, we train for 200 epochs when fitting the \textit{ground-truth} data, which is necessary due to the smaller number of ground-truth student responses. We reserve a random validation set of 20\% of all responses to determine how predictive our ground-truth IRT parameters are, comparing the most likely student scores predicted by the IRT model and the ground-truth scores in the validation set. For Smarter Balanced, we measure performance using QWK since scores are ordinal, with the IRT model achieving a QWK of $0.57$ on the validation set. For CodeWorkout, we measure performance using PCC since scores are continuous, with the IRT model achieving a PCC of $0.49$. These values show moderate predictability, indicating that while the student scores are difficult to predict, the parameters are well-calibrated.

% SUBSECTION

\section{Additional Dataset Details}
\label{sec:appendix-dataset}

\subsection{Cross-Validation Splitting}

We create our cross-validation splits such that 1) each item appears in train, validation, and test at least once, and 2) the distribution of item difficulties should be roughly the same across train, validation, and test over all folds. We achieve these requirements using the following algorithm. First, we sort all items according to their difficulty. Second, we partition all items into 10 difficulty buckets of up to 5 items each. Third, we sequentially construct a list of items by adding one random item at a time from each bucket in a round-robin fashion; the resulting list is striped by difficulty bucket. Finally, we evenly rotate this list 5 times to form the 5 cross-validation folds, partitioning each rotation at 60\% and 80\% to create a train/validation/test split.

\subsection{Passage Summarization}

Several Smarter Balanced questions include long reading passages, up to $4,000$ tokens in length, leading to increased memory usage and training time; therefore, we prompt \texttt{Llama-3.1-70B-Instruct}, loaded with NF4 quantization, to summarize these passages while retaining the main elements and original writing style. We show this prompt in Appendix~\ref{sec:prompts}.

\section{Data Scaling Experiment}

\begin{figure}[h]
    \centering
    \includegraphics[width=\linewidth]{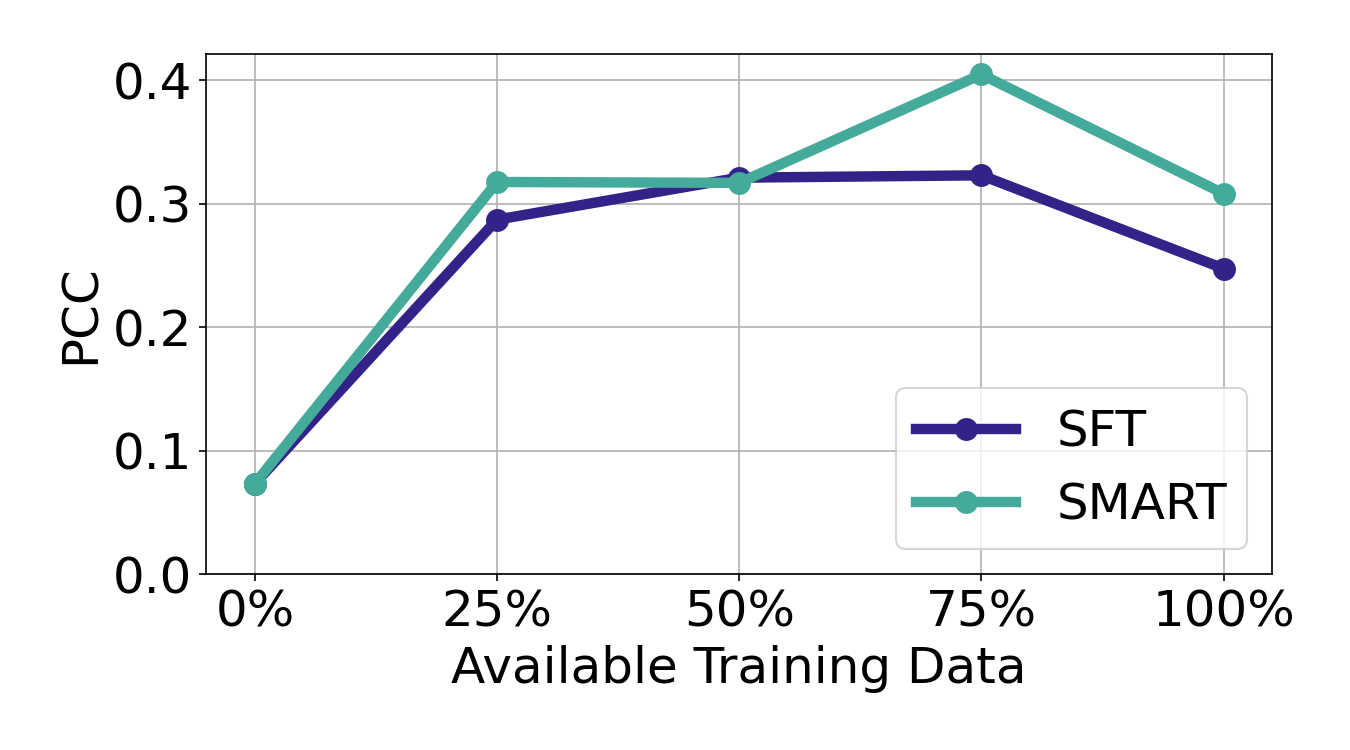}
    \caption{Results of data scaling experiment. Even a small amount of data is sufficient to achieve high performance. SMART begins to outperform SFT more significantly when more data is available.}
    \label{fig:scaling}
\end{figure}

We conduct an experiment where we vary the amount of data available for training by using random subsets of the student responses in the training set at 0\%, 25\%, 50\%, 75\%, and 100\%. We re-train both SFT and SMART using the 1B model on these subsets, and report the PCC averaged over all cross-validation test splits on CodeWorkout. To only examine the impact of training data on the student model, we use the scoring model trained on all data, since in reality a stronger scoring model may be available (such as one that runs test cases on generated code). 

We show the results of this data scaling experiment in Figure~\ref{fig:scaling}. We observe that even a small amount of training data is sufficient to drastically increase performance from zero-shot (25\% of the training data contains 1,636 responses and 3,027 preference pairs in total). SFT and SMART perform similarly until 75\%, when SMART begins to outperform SFT. Surprisingly, performance drops for both methods from 75\% to 100\%, indicating that some training responses may negatively impact performance, such as ones that represent item types that are not present in the test set. This result further reflects the challenges of testing on a small number of items.

\section{Statement on Scientific Artifacts}

\paragraph{Software and Models}
To the best of our knowledge, all software and models we build our implementation on have open-source licenses or no available license. Additionally, we are within their intended terms of use, specifically, scientific research. If we release code, we will ensure the license and terms reflect the sources we build on.

\paragraph{Dataset}
The Smarter Balanced dataset contains proprietary questions and private student data. We do not examine the data for offensive content or identifiable information, because 1) the data is private and securely stored, 2) all models trained on the data are securely stored and will not be released, and 3) the data is very large and would be difficult to effectively scan for such issues. We do not check for such issues on CodeWorkout either, because it is a public and widely used dataset, and such investigations are beyond the scope of this work.

% SECTION

\clearpage

\onecolumn

\section{Qualitative Examples}

\begin{table*}[ht]
\small
\centering
\begin{tabular}{p{0.14\linewidth}p{0.76\linewidth}}

\toprule

Passage & Source \#1 - Meet the Robots

Robots are not just a thing of the future; they are already doing many jobs today...\newline

Source \#2 - Robots That Play Well with Others

Robots are not just for work; they can also play. Some robots can sing, dance, and even play music...\newline

Source \#3 - When the Car Is the Driver

The state of Nevada has finalized new rules that will make it possible for robotic self-driving cars to receive their own special driving permits...
\\

\midrule

Question & Many robots are designed to do normal tasks that improve people's lives or jobs. Provide two pieces of evidence from different sources that support this idea and explain how each example supports the idea.\\

\midrule
Ability & Response\\

\midrule
\rowcolor{gray!21} \multicolumn{2}{c}{SFT}\\
\midrule
Low & The reason why these are good robots is because they can help with anything that you need. They can help with food, water, and other things that you need. They can also help with work, like doctors, and other things that you need to do.\\
Medium & In source 1 it talks about a robot named "Gemini Scout" that is a remote controlled robot. It can also be used for mine, farms, and doctors. It can also help people who are in danger. It can also help with rescues. In source 2 it talks about...\\
High & A few robots that are designed to do normal tasks that improve people's lives or jobs are, robots like the "Gemini-Scout" which can be used to rescue people that are in danger, such as in a mine or a building. It can also be used to test the air for dangerous gases. The "Gemini-Scout" is very useful to people because it helps them in many different ways. Another robot that is useful...\\

\midrule
\rowcolor{gray!21} \multicolumn{2}{c}{SMART (ours)}\\
\midrule
Low & The reason why i think the robots are good is becuase they can help and they are smart.\\
Medium & In source 1 it talks about a robot named "Gemini Scout" that can carry food and water. It can also collect information about the air that comes out of the vents in the volcano. In source 2 it talks about...\\
High & A large number of robots are designed to help people with tasks that make their lives easier and to improve job efficiency. A piece of evidence from "Meet the Robots" is a robot that is used in mines to help search for lost or injured people. The robot is called the Gemini-Scout, and it is a remote-controlled robot. This supports the idea because it is helping to rescue people who are injured or missing...\\

\bottomrule
\end{tabular}
%\vspace{-.2cm}
\caption{Examples of student responses generated by SFT and SMART for a publicly available Smarter Balanced item, conditioned on abilities in the low, medium, and high ranges. SMART adapts better to the instructed ability than SFT. For low ability, SMART introduces a spelling error and poor capitalization, and does not back up its claim, while SFT does not answer the question correctly but uses better sentence structure. For medium ability, both methods provide evidence but do not explain how they support the claim. For high ability, SMART produces structured writing, provides evidence, and explains how its evidence supports the claim, while SFT is less explicit about how its evidence supports the claim.}

\label{tab:qual_ex_sb}
%\vspace{-.2cm}
\end{table*}

\begin{table*}
\small
\centering
\begin{tabular}{p{0.14\linewidth}p{0.76\linewidth}}

\toprule

Passage & Plants need food to survive, just as animals do; but, unlike animals, they do not eat solid foods. Instead, plants get most of the food they need from the sun through a process called photosynthesis. Using this process, plants can absorb the sun's light energy to make their own food. Although they cannot move around to find food, plants can make their food anywhere as long as they have three things: carbon dioxide, water, and light...
\\

\midrule

Question & Write an introduction to the report that establishes and introduces a clear claim about how plants make their own food.\\

\midrule
Ability & Response\\

\midrule
\rowcolor{gray!21} \multicolumn{2}{c}{SFT}\\
\midrule
Low & Photosynthesis is the process by which plants make their own food from the sun.\\
Medium & Plants are very important to us because they give us oxygen, that we need to survive. Plants are very important because they give us oxygen.\\
High & Plants are very important to us. We rely on them for food and for oxygen. Plants make all of this possible. They take the carbon dioxide, the water, and the light, and turn them into sugar and oxygen. That is why photosynthesis is so important to animals.\\

\midrule
\rowcolor{gray!21} \multicolumn{2}{c}{SMART (ours)}\\
\midrule
Low & that the plants are good for us and the world\\
Medium & Plants are one of the most important things on earth, without them we would not be alive today. Plants are very important to us humans because without them we wouldn't be able to survive, they also help us in many ways like we can make food out of them and plants help us by providing oxygen.\\
High & Photosynthesis is the process by which plants create food. The process of photosynthesis is very important to all living things, especially because without it, the air we breathe would be filled with only carbon dioxide, and without oxygen, animals and humans would not be able to survive. In this report, you will learn more about how plants make their own food.\\

\bottomrule
\end{tabular}
%\vspace{-.2cm}
\caption{Examples of student responses generated by SFT and SMART for a publicly available Smarter Balanced item, conditioned on abilities in the low, medium, and high ranges. SMART adapts better to the instructed ability than SFT. For low ability, SMART uses very simple language and doesn't discuss how plants make food, while SFT uses more formal language and does mention how plans make food, which would be atypical for a low ability student. For medium ability, both methods write a coherent introduction but don't mention how plants make food; SMART writes a slightly longer response. For high ability, SMART makes sure to motivate the importance of photosynthesis as well as directly answer the question by stating how plants make food, while SFT neglects to mention how plants make food.}

\label{tab:qual_ex_sb_2}
%\vspace{-.2cm}
\end{table*}

\pagebreak

\small{
\begin{longtable}[ht]{p{0.14\linewidth}p{0.66\linewidth}p{0.1\linewidth}}
% \small
% \centering
% \begin{tabular}{p{0.14\linewidth}p{0.66\linewidth}p{0.1\linewidth}}

\toprule
Question & Return the centered average of an array of ints, which we'll say is the mean average of the values, except ignoring the largest and smallest values in the array. If there are multiple copies of the smallest value, ignore just one copy, and likewise for the largest value. Use int division to produce the final average. You may assume that the array has a length of 3 or more.\\

\midrule
Ability & Response & Score\\

\midrule
\rowcolor{gray!21} \multicolumn{3}{c}{Human Student}\\
\midrule
Low & \begin{footnotesize} \begin{verbatim}public int centeredAverage(int[] nums)
{
    int merlin = 0;
    int less = nums[0];
    int more = nums[0];
    
    for (int i = 0; i < nums.length; i++)
    {
        merlin = merlin + nums[i];
        less = Math.min(less, nums[i]);
        more = Math.max(more, nums[i]);
    }
    return (sum - (min-max)/nums.length);   
}\end{verbatim} \end{footnotesize} & 0.0\\
Medium & \begin{footnotesize} \begin{verbatim}public int centeredAverage(int[] nums)
{
    int max = 0;
    int min = nums[0];
    
    for (int i = 0; i < nums.length; i++)
    {
        if (nums[i] > max)
        {
            max = nums[i];
        }
        
        else if (nums[i] < min)
        {
            min = nums[i];
        }
    }
    
    int sum = 0;
    int ct = 0;
    
    for (int n : nums)
    {
        if (n != min && n != max)
        {
            sum += n;
            ct++;
        }
    }
    return sum/ct;
}\end{verbatim} \end{footnotesize} & 0.67\\
High & \begin{footnotesize} \begin{verbatim}public int centeredAverage(int[] nums)
{
    int sum = 0;
    int min = nums[0];
    int max = nums[0];
         
    for(int i = 0; i < nums.length; i++) {
        sum += nums[i];
        min = Math.min(min, nums[i]);
        max = Math.max(max, nums[i]);
    }
                           
    return (sum - min - max) / (nums.length - 2);
}\end{verbatim} \end{footnotesize} & 1.0\\

\midrule
\rowcolor{gray!21} \multicolumn{3}{c}{SFT}\\
\midrule
Low & \begin{footnotesize} \begin{verbatim}public int centeredAverage(int[] nums)
{
    int sum = 0;
    for (int i = 0; i < nums.length - 2; i++)
    {
        sum = sum + nums[i];
    }
    return sum / (nums.length - 2);
}\end{verbatim} \end{footnotesize} & 0.68\\
Medium & \begin{footnotesize} \begin{verbatim}public int centeredAverage(int[] nums)
{
    int sum = 0;
    for (int i = 0; i < nums.length; i++)
    {
        sum = sum + nums[i];
    }
    int avg = sum / (nums.length - 2);
    return avg;
}\end{verbatim} \end{footnotesize} & 0.75\\
High & \begin{footnotesize} \begin{verbatim}public int centeredAverage(int[] nums)
{
    int sum = 0;
    
    for (int i = 0; i < nums.length; i++) {
        if (nums[i]!= Integer.MIN_VALUE && nums[i]!= Integer.MAX_VALUE)
            sum += nums[i];
    }
    
    return sum / (nums.length - 2);
}\end{verbatim} \end{footnotesize} & 0.73\\

\midrule
\rowcolor{gray!21} \multicolumn{3}{c}{SMART (ours)}\\
\midrule
Low & \begin{footnotesize} \begin{verbatim}public int centeredAverage(int[] nums)
{
    int i = 0;
    int sum = 0;
    for (int i = 0; i < nums.length; i++)
    {
        sum = sum + nums[i];
    }
    return sum / nums.length - 1;
}\end{verbatim} \end{footnotesize} & 0.27\\
Medium & \begin{footnotesize} \begin{verbatim}public int centeredAverage(int[] nums)
{
    int sum = 0;
    int min = Integer.MAX_VALUE;
    int max = Integer.MIN_VALUE;
    
    for (int i = 0; i < nums.length; i++) {
        if (nums[i] > max) {
            max = nums[i];
        }
        if (nums[i] < min) {
            min = nums[i];
        }
        sum += nums[i];
    }
    
    return (sum - max - min) / (nums.length - 2);
}\end{verbatim} \end{footnotesize} & 0.94\\
High & \begin{footnotesize} \begin{verbatim}public int centeredAverage(int[] nums) {
  int min = Integer.MAX_VALUE;
  int max = Integer.MIN_VALUE;
  int sum = 0;
  
  for(int num : nums) {
    if(num < min) min = num;
    if(num > max) max = num;
    sum += num;
  }
  
  sum -= min + max;
  return sum / (nums.length - 2);
}\end{verbatim} \end{footnotesize} & 0.98\\

\bottomrule
% \end{tabular}
%\vspace{-.2cm}
\caption{Examples of real student responses and responses generated by SFT and SMART for a CodeWorkout item, conditioned on abilities in the low, medium, and high ranges. Scores for real students represent the portion of test cases that passed; scores for simulated students are assigned by our trained scoring model. The true difficulty for the question is -0.57, the SFT difficulty prediction is 0.22, and the SMART difficulty prediction is -0.41. SMART responses are much more diverse than SFT responses, likely resulting in the more accurate prediction. To further demonstrate this, we report the average score for each ability quartile from each source, where real students and SMART show a gradual increase in score with ability, while SFT does not show any distinction between abilities. Real students: [0.22, 0.53, 0.71, 0.95], SFT: [0.81, 0.79, 0.82, 0.83], SMART: [0.57, 0.79, 0.85, 0.93].}

\label{tab:qual_ex_codeworkout}
%\vspace{-.2cm}
\end{longtable}
}

\normalsize

\twocolumn

\section{Prompts}
\label{sec:prompts}

We show our prompts below for the simulated student model (Table~\ref{tab:prompt-sim-stud}), the automated scoring model (Table~\ref{tab:prompt-score}), ICL for difficulty prediction (Table~\ref{tab:prompt-icl}), and summarizing reading passages (Table~\ref{tab:prompt-passage}). We note that because the simulated student and automated scoring models are finetuned, their system prompts are much less detailed. Doing so reduces GPU memory, and is permissible since the models are able to learn to perform their tasks from the data, rather than relying on instructions in the prompt.

\begin{table*}[ht]
    \centering
    \small
    \begin{tabular}{p{2cm}p{12cm}}
        \toprule
        \textbf{System} & You are a student of \{ability\_level\} ability responding to a stimulus. \\
        \midrule
        \textbf{User} & Given the following **Stimulus**, give a response to the **question**:\newline

**Stimulus**\newline

--- \textit{for Smarter Balanced} ---\newline
Passage:\newline
\{passage\}\newline
------\newline

Question:\newline
\{question\}\newline

--- \textit{for CodeWorkout} ---\newline
Reference Solutions:\newline
\{reference solutions\}\newline
------\\
        \bottomrule
    \end{tabular}
    \caption{Prompt for the simulated student model.}
    \label{tab:prompt-sim-stud}
\end{table*}

\begin{table*}[ht]
    \centering
    \small
    \begin{tabular}{p{2cm}p{12cm}}
        \toprule
        \textbf{System} & You are a teacher scoring student responses to open-ended questions. \\
        \midrule
        \textbf{User} & Score the student response using the rubric to the question by only outputting a single integer between 0 and \{max\_score\} inclusive.\newline

--- \textit{for Smarter Balanced} ---\newline
Rubric:\newline
\{rubric\}\newline

Passage:\newline
\{passage\}\newline
------\newline

Question:\newline
\{question\}\newline

--- \textit{for CodeWorkout} ---\newline
Reference Solutions:\newline
\{reference solutions\}\newline
------\newline

Student response: \{response\}\newline

Score:\\
        \bottomrule
    \end{tabular}
    \caption{Prompt for the automated scoring model.}
    \label{tab:prompt-score}
\end{table*}

\begin{table*}[ht]
    \centering
    \small
    \begin{tabular}{p{2cm}p{12cm}}
        \toprule
        \textbf{System} & You are an experienced English teacher. Your job is to predict the difficulty of a reading comprehension short answer question. Please follow these instructions carefully:\newline
- You will be given a question and optionally an associated passage. Use both of these to predict the question's difficulty.\newline
- You will also be given several example questions and their ground-truth difficulties. Use these to calibrate your prediction for the current question.\newline
- Difficulties are continuous real numbers, and can be positive or negative. A higher value means the question is more difficult.\newline
- First briefly think step-by-step about the question's difficulty based on its content and the examples.\newline
- Write your output with the following template: ``Thinking: <step-by-step reasoning>\textbackslash nDifficulty: <real-valued difficulty prediction>''. Ensure that you only output a number for the difficulty prediction. \\
        \midrule
        \textbf{User} & \#\#\#\# Example 1 \#\#\#\#\newline

--- \textit{for Smarter Balanced} ---\newline
Passage:\newline
\{passage\}\newline
------\newline

Question:\newline
\{question\}\newline

--- \textit{for CodeWorkout} ---\newline
Reference Solutions:\newline
\{reference solutions\}\newline
------\newline

Difficulty: \{difficulty\}\newline

\#\#\#\# Example 2 \#\#\#\#\newline
...\newline

\#\#\#\# Example 3 \#\#\#\#\newline
...\newline

\#\#\#\# Current Question \#\#\#\#\newline

--- \textit{for Smarter Balanced} ---\newline
Passage:\newline
\{passage\}\newline
------\newline

Question:\newline
\{question\}\newline

--- \textit{for CodeWorkout} ---\newline
Reference Solutions:\newline
\{reference solutions\}\newline
------\\
        \bottomrule
    \end{tabular}
    \caption{Prompt for in-context learning difficulty prediction.}
    \label{tab:prompt-icl}
\end{table*}

\begin{table*}[ht]
    \centering
    \small
    \begin{tabular}{p{2cm}p{12cm}}
        \toprule
        \textbf{System} & You are an English teacher. You will be given a reading passage. Your job is to rewrite the passage to make it shorter. Please follow these instructions carefully:\newline
- Your version of the passage should be significantly shorter.\newline
- It should contain all of the main ideas and plot elements of the original passage.\newline
- Do not change the writing style of the passage.\newline
- If the passage contains multiple texts/sources, rewrite all of them in independent paragraphs.\newline
- Do not write an answer to the question.\newline
- Your output will be used as the new passage. Therefore, only output the rewritten passage. Do not write any explanation of your output at all. \\
        \midrule
        \textbf{User} & \{passage\}\\
        \bottomrule
    \end{tabular}
    \caption{Prompt for passage summarization.}
    \label{tab:prompt-passage}
\end{table*}

\end{document}